\documentclass[conference]{IEEEtran}

\usepackage[letterpaper, left=1in, right=1in, bottom=1in, top=0.75in]{geometry}
\usepackage{cite}
\usepackage{amsmath,amssymb,amsfonts}
\usepackage{textcomp}

\usepackage{url}
\usepackage{bm}
\usepackage{multirow}
\usepackage{tabularx}
\usepackage{hyperref}
\usepackage{soul}
\usepackage{booktabs}
\usepackage{times}
\usepackage{epsfig}
\usepackage{graphicx}
\usepackage{authblk}
\usepackage{multirow}
\usepackage{tabularx}
\usepackage{balance}
\usepackage{subfigure}
\usepackage{balance}
\usepackage{xcolor}
\usepackage{color, colortbl}
\usepackage[noend]{algpseudocode}
\usepackage{enumitem}
\setlist{nosep}
\usepackage{diagbox}
\usepackage{etoolbox}
\usepackage{makecell}

\usepackage{academicons}

\makeatletter
\patchcmd{\@makecaption}
  {\scshape}
  {}
  {}
  {}
\makeatother

\def\BibTeX{{\rm B\kern-.05em{\sc i\kern-.025em b}\kern-.08em
    T\kern-.1667em\lower.7ex\hbox{E}\kern-.125emX}}

\newcommand{\newcommentAdd}[1]{\textcolor{black}{#1}}
\newcommand{\newcommentDel}[1]{\textcolor{black}{#1}}

\usepackage{pifont}
\newcommand{\xmark}{\ding{55}}%

\def\eg{\emph{e.g}.} 
\def\ie{\emph{i.e}.}

\def\etal{\emph{et~al}.}

\DeclareMathOperator*{\argmax}{argmax} 

\definecolor{lightgray}{gray}{0.9}
\begin{document}
\Urlmuskip=0mu plus 1mu
\title{Multilingual Communication System with \\Deaf Individuals Utilizing Natural and \\Visual Languages}

\author[1,2]{Tuan-Luc Huynh$^\dagger$\thanks{$\dagger$These authors have equal contributions}}
\author[1,2]{Khoi-Nguyen Nguyen-Ngoc$^\dagger$}
\author[1,2]{Chi-Bien Chu$^\dagger$}
\author[1,2]{\\Minh-Triet Tran}
\author[1,2]{Trung-Nghia Le$^\ddagger$\thanks{$\ddagger$Corresponding author}}

\affil[1]{Faculty of Information Technology, University of Science, VNU-HCMC, Vietnam}
\affil[2]{Vietnam National University, Ho Chi Minh City, Vietnam}

\maketitle

\begin{abstract}
According to the World Federation of the Deaf, more than two hundred sign languages exist. Therefore, it is challenging to understand deaf individuals, even proficient sign language users, resulting in a barrier between the deaf community and the rest of society. To bridge this language barrier, we propose a novel multilingual communication system, namely MUGCAT, to improve the communication efficiency of sign language users. By converting recognized specific hand gestures into expressive pictures, which is universal usage and language independence, our MUGCAT system significantly helps deaf people convey their thoughts. To overcome the limitation of sign language usage, which is mostly impossible to translate into complete sentences for ordinary people, we propose to reconstruct meaningful sentences from the incomplete translation of sign language. We also measure the semantic similarity of generated sentences with fragmented recognized hand gestures to keep the original meaning. Experimental results show that the proposed system can work in a real-time manner and synthesize exquisite stunning illustrations and meaningful sentences from a few hand gestures of sign language. This proves that our MUGCAT has promising potential in assisting deaf communication.
\end{abstract}


\section{Introduction}
\label{sec:introduction}

Communication with deaf people is mainly based on sign language, a combination of hand gestures, facial expressions, and postures to convey semantic information. However, these visual communication systems are difficult to learn and remember, leading to barriers between the deaf community and the rest of society; this problem has not been fully solved until now.

Although technologies have been developed to understand the behaviors of deaf people, such as sign language translation via cameras and sensory gloves, they still have several issues. Sign language translation via camera systems~\cite{Hao-ICCV2021, Min-ICCV2021} needs a fixed camera and a simple background to recognize sign gestures accurately. Besides that, translating sign languages to human-understandable languages often leads to unnatural results. It causes difficulties in jobs that require smoothness in words, such as explaining new concepts or telling stories. The birth of sensory devices like smart groves~\cite{Frederic-2012, UWnews-2016} is a big step forward in this field. However, it still does not solve the problem of unnatural translations. In addition, the more modern sensors will come with high prices, which makes it difficult to reach the deaf community.



Combined with natural language, which can be expressed as text or voice, visual language can reform the communication between ordinary people and deaf/dumb people. Indeed, visual cues (\eg, images, videos, 3D models) are the best aid to express new concepts intuitively. Visual cues play an important role in deaf communication, especially in literacy education for deaf children. Visual communication efficiently bridges ordinary people with the deaf community, regardless of different nationalities or different languages.

To assist communication with deaf individuals, we propose a \textbf{MU}ltilin\textbf{G}ual \textbf{C}ommunic\textbf{AT}ion system (MUGCAT). Inspired by the adage "A picture is worth a thousand words," our system supports \newcommentDel{\st{in}} diverse cues, such as sign languages, natural languages, and visual languages, to help deaf people express their thoughts more clearly. The proposed MUGCAT system consists of two main phases: converting sign language to intermediate language, which ordinary people can understand, and enriching the translated information by reconstructing a meaningful sentence aided by illustrations. We first recognize and translate these hand gestures of deaf individuals (\ie, sign language) into human-understandable language (\ie, textual words or phases). Illustrations are then synthesized via a text-to-image model for visual communication. By transforming sign languages into pictures - universal mediums of expressiveness - our system significantly help deaf individuals convey their thoughts. Due to the limitation of sign languages, it is challenging to translate hand gestures to complete meaningful sentences for ordinary people. Therefore, we propose using an image captioning method to assist the incompleted translated text. Furthermore, our MUGCAT system can measure the semantic similarity of generated image captions with the intermediate translated text to keep the original meaning of the sign language. In this way, our MUGCAT system can help to express the intentions of deaf communicators more intuitively and clearly.

Experimental results on the WLASL dataset~\cite{Li-2020} show the potential of our MUGCAT system in assisting natural communication with deaf individuals. The proposed system can recognize sign gestures with an accuracy of $46.8\%$ in \newcommentAdd{real time}. In addition, meaning sentences for humans are generated with corresponding exquisite, beautiful, and stunning illustrations. We expect our MUGCAT system to benefit both the deaf community and the sign language research community.

Our main contributions are summarized as follows: 
\begin{itemize}

    \item We propose a novel system, namely MUGCAT, to support multilingual communication for deaf individuals. Our simple yet efficient system utilizes both natural and visual languages to enhance the interpretation of deaf communicators.
    
    \item Our MUGCAT system accurately recognizes and translates sign languages to human-understandable text. The proposed system also can transform the translated text into illustrative and expressive images in real-time performance.
    
    \item The synthesized images might be misleading; hence, we propose to use an image captioning model to select the image that best fit the translated text\newcommentAdd{, further improving the efficiency of MUGCAT}.
\end{itemize}





\section{Related Work}
\label{sec:related_work}

\subsection{Deaf Communication}


Communicating with deaf individuals mainly occurs through auditory (\eg, lip reading) and visual (\eg, sign language) modes. However, sign language is more popular than lip reading because understanding speech by visually interpreting the movements of the lips, face, and tongue is extremely challenging, even for deaf people.

With the development of modern technology, many technological devices have been invented to translate sign language into text, speech, etc. Some special sensors were made to detect hand movements. The translation glove products (EnableTalk~\cite{Frederic-2012} and SignAloud~\cite{UWnews-2016}) work on the integration of sensors that attach to the finger to record hand posture and movements and then convert sensor signals into speech through an independent processing unit. However, these products are difficult to access widely due to their high cost and difficulty to use in daily life.

\subsection{Sign Language Recognition}





Recently, many computer vision algorithms have been proposed to recognize sign language from video only, thus avoiding the dependence on costly sensor devices. Given a video, besides RGB frames, we can also obtain other modalities of input such as image depth \cite{Akhil-2021, Jianyuan-CVPR2021} and optical flow \cite{Zhaoyang-ECCV2022, Shaojie-CVPR2022} (pixel-wise motions between consecutive video frames). For RGB input only, 3D ConvNets were widely applied \cite{Carreira-CVPR2017, Shuiwang-2013} to extract spatial-temporal information from videos. Lin~\etal~\cite{Lin-2019} inserted a Temporal Shift Module into 2D ConvNets to get an accuracy commensurate with 3D ConvNets while keeping the complexity of 2D ConvNets. Komkov~\etal~\cite{Komkov-2020} combined the learned knowledge from multiple single-modality models with mutual learning technique \cite{Zhang-CVPR2018} to obtain the best model on each input modality. 

\subsection{Text-to-Image}

In the last couple of years, text-to-image models~\cite{Samah-MM2020} have attracted big tech companies' attention and thus have received rapid and massive improvements. Classifier-free guided diffusion models have recently been shown to be highly effective at high-resolution image generation, and they have been widely used in large-scale diffusion frameworks, including GLIDE~\cite{nichol-arxiv2021}, DALL-E 2~\cite{ramesh-arxiv2022}, and Imagen~\cite{meng-arxiv2022}. Nevertheless, the latest development of diffusion model-based text-to-image model, namely Stable Diffusion~\cite{rombach-CVPR2022}, has been the most significant impact since its release. Stable Diffusion offers excellent image quality while significantly lowering the computation cost. What makes Stable Diffusion exceptionally attractive compared to other competitors due to its open-source. On the other hand, Google and OpenAI do not intend to open-source Imagen~\cite{imagen} and DALL-E 2~\cite{ramesh-arxiv2022}, respectively.

While the artificial intelligence community has dominantly used text-to-image models to create beautiful artworks, there is little attention on using these models on real-world problems. In this work, we customized Stable Diffusion to generate meaningful and expressive images from the translated sign language text in a real-time manner to help visualize conversation with or between sign language users.

\subsection{Image Captioning}

Research on image captioning in recent years generally uses the encoder-decoder architecture. The encoder extracts the visual information from images for the decoder, which generates an acceptable description. In the early, the encoder was a CNN backbone \cite{Xu-ICML2015, Rennie-CVPR2017}. Later, it was replaced by an object detector such as Faster R-CNN to extract object-level features~\cite{Anderson-CVPR2018}. This proved more efficient and improved performance because the object information and their relationships are very useful in describing an image. However, due to the high computational cost of the object detection model, it is hard to apply in a problem that requires high speed, such as communication. Besides that, Transformer applications in the encoder to extract features or the decoder for caption generating task \cite{Li-ICCV2019, Pan-CVPR2020} also demonstrated surprising efficiency improvements. 

In this study, with the aim of balancing accuracy and efficiency, we used a recent state-of-the-art method~\cite{Nguyen-arxiv2022} which proposed a Transformer-only neural architecture utilizing dual visual features to improve performance and increase speed. 


\section{Proposed System}
\label{sec:proposed_system}
\subsection{Overview}

\begin{figure}[t]
  \centering
  \includegraphics[width=1\linewidth]{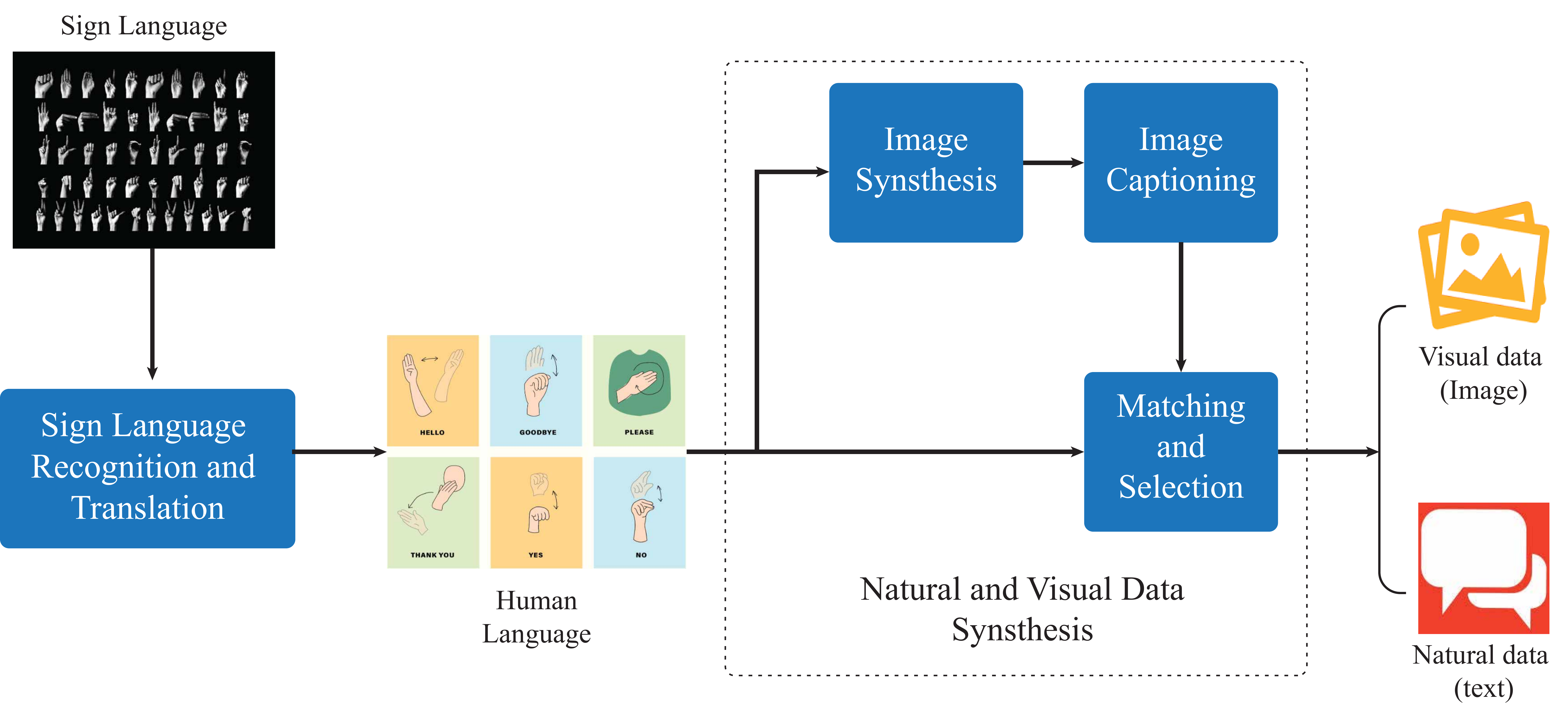}
  \caption{Pipeline of our multilingual communication (MUGCAT) system}
  \label{fig:pipeline}
\end{figure}

Figure~\ref{fig:pipeline} illustrates the pipeline of our proposed multilingual communication (MUGCAT) system, which consists of two main components: Sign language recognition and translation (SLRT), Natural and visual data synthesis. First, words obtained from SLRT are illustrated by the text-to-image module resulting in several images. Then, image captioning is carried out to achieve complete descriptions of all synthesized images. Finally, with each image and its description, we compare its semantic similarity with the translated keywords from SLRT to choose the most suitable image and description. Unlike conventional SLRT systems that need to correctly recognize the whole sentence to express the meaning, our system generates a suggested image and complete description to represent the keywords made in sign language to overcome the disadvantage of missed recognized sign language.

\subsection{Sign Language Recognition and Translation}

Sign language recognition aims to predict the sequence of signs performed in a video, while sign language translation further translates the signs into spoken/written languages. To synthesize images that fully convey the meaning of a signer, we only need to identify some keywords from the hand gesture sequence. Therefore, the recognition task is more suitable for our MUGCAT system because it is simple but still responsive to the system. Due to the lack of suitable datasets in this domain, we treat the problem as an action recognition task, where the objective is to identify single words from short clips. This simplifies the problem and meets our requirement for indicating visual keywords. 

In this work, we employed and compared several action recognition methods \cite{Carreira-CVPR2017, Lin-2019, Komkov-2020} on WLASL dataset~\cite{Li-2020}, the largest video dataset of word-level American sign language (ASL). The main ideas of employed methods are summarized as follows:

\textit{Two-Stream Inflated 3D ConvNets (I3D)}~\cite{Carreira-CVPR2017} combines two 3D ConvNets (one for RGB image stream, one for optical-flow stream) to both take advantage of pre-trained ImageNet weights and force the model to learn motion features directly. Two 3D ConvNets are trained separately, and their predictions are averaged at test time. 

\textit{Temporal Shift Module (TSM)}~\cite{Lin-2019} inserted TSM module into 2D ConvNets to capture temporal relationships between video frames. Feature maps are shifted along the temporal dimension to maintain 2D ConvNet's complexity while achieving the performance of 3D ConvNets.

\textit{Mutual Modality Learning (MML)}~\cite{Komkov-2020} ensembled knowledge from single-modality models into a single model to obtain the best single-modality model for each modality. The algorithm can be summarized in three steps: train two separate networks $A_1$, $A_2$ on the RGB modality; respectively initialize two networks $B_1$, $B_2$ with the weights of $A_1$, $A_2$, then train $B_1$, $B_2$ together using mutual learning technique on RGB modality; from $B_1$'s weights, initialize $N$ models $C_1$, $C_2$, ..., $C_N$ corresponding to $N$ different modalities (RGB, optical flow, and depth), then train these $N$ models together using mutual learning.

\subsection{Natural and Visual Data Synthesis}

\subsubsection{Text-To-Image Synthesis}

Stable Diffusion~\cite{rombach-CVPR2022}, a state-of-the-art diffusion-based text-to-image model, is the core component of our system, which strives to actualize the adage "A picture is worth a thousand words." This method can offer excellent image quality while significantly lowering computation costs. 

However, the sequential sampling process of diffusion-based models is time-consuming. As a result, the text-to-image module is also the bottleneck of our system. To overcome this limitation, we customized hyperparameters of Stable Diffusion to retain high-quality images while significantly reducing the sampling process time.

Another issue that affects our system performance is the relevancy of synthesized images. Prompt engineering (\ie, prompt modifiers) is necessary for guiding the text-to-image models to generate superior-quality art. However, in our proposed system, the prompt text for Stable Diffusion is limited keywords from the SLRT module. Therefore, it is unavoidable that the prompt's quality is limited, which leads to potential drops in generated image relevancy. We addressed this issue by introducing the image captioning model in the system's next stage, which serves as a filter to select the most relevant image.

\subsubsection{Image Captioning}

Image captioning methods are classified into two main approaches: grid features and region features. Methods based on grid features directly extract object features from high-layer feature maps of the whole image. Thus, generated captions can contain information about the whole image. Meanwhile, methods based on region features~\cite{Anderson-CVPR2018} rely on detecting objects in the image and then extracting local features of image regions to infer results. However, detected objects cannot represent the overall context of the image nor the relationships of objects that affect generated captions. 

In this work, we used Grid- and Region-based Image captioning Transformer (GRIT)~\cite{Nguyen-arxiv2022}, a state-of-the-art image captioning method, which uses both types of mentioned features to enhance both contextual information and object-level information. Grid features are extracted using a standard self-attention Transformer, and region features are extracted by Deformable DETR detector~\cite{Zhu-ICLR2021}. Then, the extracted features are fed to a caption generator based on Transformer to generate the final caption. In this step, we employed  Parallel Cross-Attention~\cite{Nguyen-arxiv2022} to \newcommentAdd{relate} between dual visual features and caption words.

\subsubsection{Matching and Selection}

SLRT generates incomplete keywords; thus, the images synthesized from the previous step inevitably are not completely consistent with each other and the communicator's expression. Therefore, we propose an extra step of matching and selecting the caption whose meaning is closest to the input keywords. From there, our system is able to recover the complete sentence that the communicator wanted to express from just the discrete words.

Concretely, given $K$ results $\{I_1, I_2,\dots, I_K\}$ of the previous step, the goal of the matching and selection is to find the most suitable pair of image $\widehat{I}$ and its caption $q_{\widehat{I}}$. \newcommentAdd{For} each caption sentence, we measure its semantic similarity with keywords obtained from SLRT. We used Sentence Transformers~\cite{Reimers-arxiv2019}, denoted by $\psi(\cdot)$, to compute sentence embeddings and evaluate them with cosine similarity. Mathematically, it performs a maximization expressed as:

\begin{equation}
\label{eq:similarity}
    \{\widehat{I}, q_{\widehat{I}}\} = \argmax_{i \in \{1, 2,\dots,K\}}D(\psi(q_{I_i}), \psi(Q)),
\end{equation}
where $Q$ is a sentence that includes keywords received from SLRT, $D(\cdot)$ is a cosine similarity function.

\section{Experiments}
\label{sec:experiments}

In this section, we elaborate on the extensive experiments conducted on our proposed system. All experiments were tested on a machine with a single Nvidia V100 GPU.

\subsection{Sign Language Recognition}
\label{sec:exp-SLR}

\begin{table}[t]
    \centering
    \label{tab:compare-SLR-methods}
    \caption{Accurary and efficiency on the WLASL \cite{Li-2020} test set. All the compared methods utilize a pre-trained backbone on ImageNet, and then they were finetuned on the WLASL dataset. The FPS was \newcommentAdd{measured} on a Nvidia V100 GPU.}
    \resizebox{\columnwidth}{!}{%
        \begin{tabular}{|c|c|c|c|c|c|}
        \hline
        \textbf{\thead{Method}} & \textbf{\thead{Pretraining \\ Dataset}} & \textbf{\thead{Accuracy} (\%)} & \textbf{\thead{FPS \\ \scalebox{.8}{(infer only)}}} & \textbf{\thead{FPS \\ \scalebox{.8}{(infer $\&$ load data)}}} \\ 
        \hline
        I3D \cite{Carreira-CVPR2017} & BSL1K \cite{Albanie-ECCV2020} & \textbf{46.8} & \textbf{1429} & 95 \\ 
        & Kinetic \cite{Carreira-CVPR2017} & 32.5 & & \\ 
        \hline
        TSM \cite{Lin-2019} & \xmark & 20.8 & 357 & 60 \\
        & Kinetic \cite{Carreira-CVPR2017} & 13.9 & & \\
        \hline
        MML \cite{Komkov-2020} & \xmark & 20.8 & 323 & \textbf{104} \\
        \hline
        \end{tabular}%
    }
\end{table}

As shown in Table~\ref{tab:compare-SLR-methods}, we compared several action classification methods \cite{Carreira-CVPR2017, Lin-2019, Komkov-2020} on the WLASL \cite{Li-2020} test set. In detail, for TSM \cite{Lin-2019}, we trained a model from scratch and fine-tuned another model, which was pre-trained on the Kinetic \cite{Carreira-CVPR2017} dataset. For MML \cite{Komkov-2020}, we trained a model from scratch with only RGB input as WLASL \cite{Li-2020} dataset does not provide optical flow or depth annotations. All the networks above use an ImageNet-pre-trained ResNet50 as the backbone. Lastly, we reused two public I3D~\cite{Carreira-CVPR2017} with different pretraining datasets~\cite{Li-2020, Albanie-ECCV2020} for our experiment.

We first evaluated  top-1 accuracy on the WLASL \cite{Li-2020} test set. The main challenge of this dataset is the number of words to classify up to 2,000, while the number of videos in the training set is just over 14,000. Therefore compared methods only achieve acceptable accuracy. I3D achieved the top-1 accuracy of $46.8\%$, which is also state-of-the-art top-1 accuracy on the WLASL test set. 

We then evaluated the efficiency of methods by measuring the execution time and the number of processed frames on the whole test set to obtain the average FPS. All methods can run in a real-time manner. Especially, MML \cite{Komkov-2020} can achieve 104 FPS counting all initialization steps, such as loading the model, preparing the dataset, etc. We also tried to compute FPS in the practice scenario, where SLRT methods directly process the video stream and ignore the initialization steps. All methods can achieve more than 300 FPS, and I3D~\cite{Carreira-CVPR2017} achieved a surprising speed of 1429 FPS. The results show that these models have the potential to be deployed on mobile devices and embedded systems while still achieving real-time speed.

\subsection{Stable Diffusion Hyperparameters Adjustment}

The default settings of Stable Diffusion~\cite{rombach-CVPR2022} hardly achieve near real-time performance. This section discusses our extensive experiments in various settings to discover the optimal trade-off point between execution time and image quality. Specifically, we focus on the number of sampling steps, the desired resolution, and the number of samples. We used the public checkpoint \textit{sd-v1-4.ckpt} in our experiments.

\begin{figure}[t]
  \centering
  \includegraphics[width=0.85\linewidth]{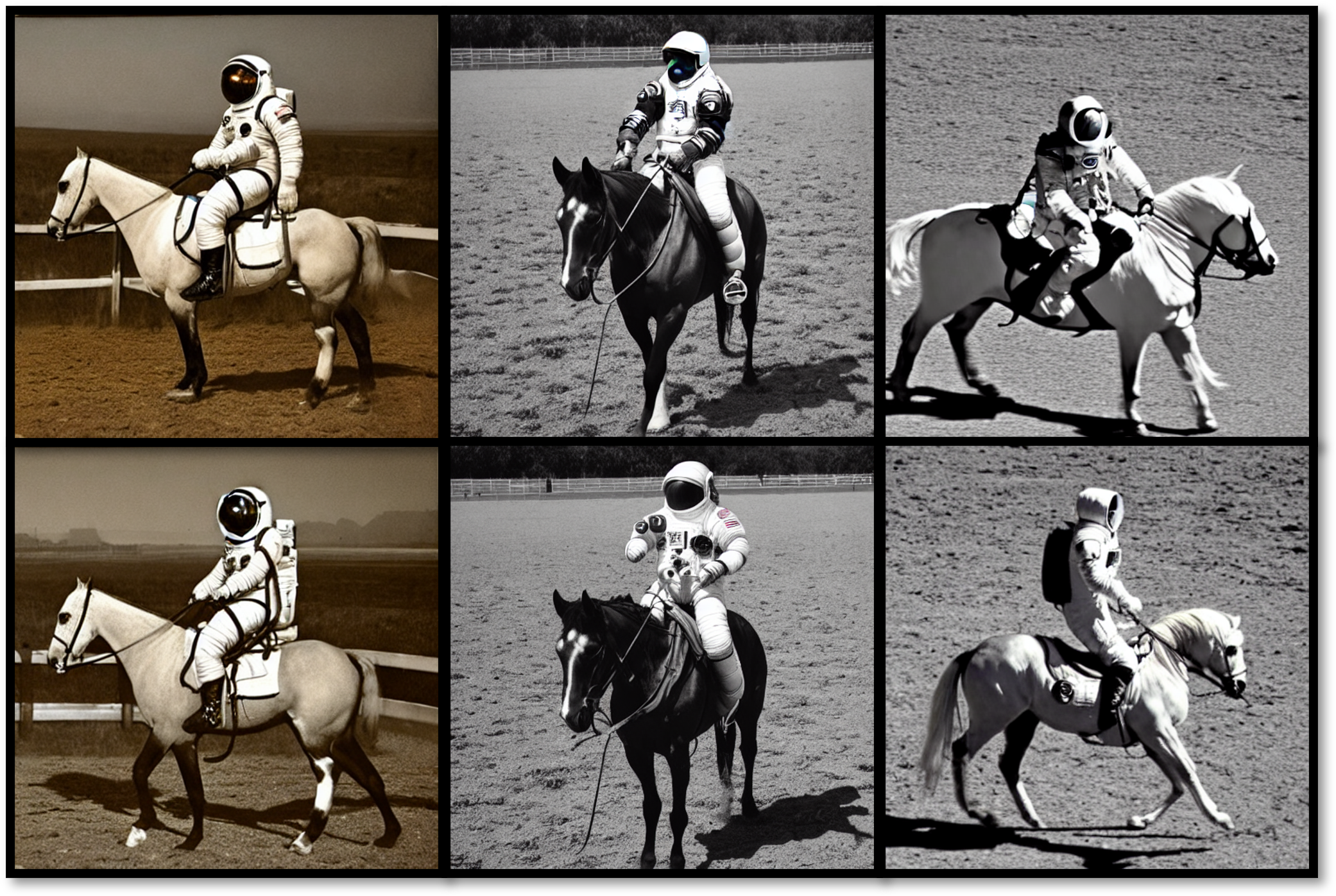}
  \caption{The renowned "a photograph of an astronaut riding a horse." The top and down \newcommentAdd{rows} are 20 and 50 sampling steps, respectively.}
  \vspace{-2mm}
  \label{fig:ddim}
\end{figure}

The number of sampling steps is the most crucial hyperparameter that directly controls the quality of generated images and positively correlates with the execution duration. The default hyperparameter of 50 sampling steps using PLMS sampler~\cite{liu-ICLR2022} generates high-quality images. Since our system ideally should work in real-time performance, we figured that 20 sampling steps could speed up 2.4 times while having subtle drops in contextual information, as demonstrated in Fig.~\ref{fig:ddim}. Indeed, Table \ref{tab:ddim_steps} shows that \newcommentAdd{setting the sampling steps to 20 is optimal with the FID of 33.5, approximately equivalent to higher sampling steps} but can process a batch of 128 images in only 15 seconds. Going lower than 20 sampling steps results in a surge of FID scores.

\begin{table}[t]
    \centering
    \label{tab:ddim_steps}
    \caption{Performance of Stable Diffusion on a single Nvidia V100 GPU. The prompt is "A beautiful flower garden \newcommentAdd{on} a sunny day with a valley background." Resolution is $512 \times 512$. The FID score~\cite{Heusel-NeurIPS2017} was calculated using 50 sampling steps as the real distribution, 128 images per distribution. The optimal hyperparameter is 20 sampling steps, which can keep the image quality but speed up 2.4 times.}
        \begin{tabular}{|l|l|l|}
        \hline
        \textbf{Sampling steps} & \textbf{FID Score $\downarrow$} & \textbf{Seconds per Batch $\downarrow$} \\ 
        \hline 50 & 0 & 35.50\\
        \hline 45 & 33.43 & 32.05\\
        \hline 40 & 30.44 & 28.66\\
        \hline 35 & 31.70 & 25.24\\
        \hline 30 & 31.55 & 21.79\\ 
        \hline 25 & 33.19 & 18.39\\ 
        \hline \rowcolor{lightgray} 20 & 33.51 & 14.97\\
        \hline 15 & 40.33 & 12.25\\
        \hline
        \end{tabular}%
\end{table}

\begin{figure}[t]
  \centering
  \includegraphics[width=1\linewidth]{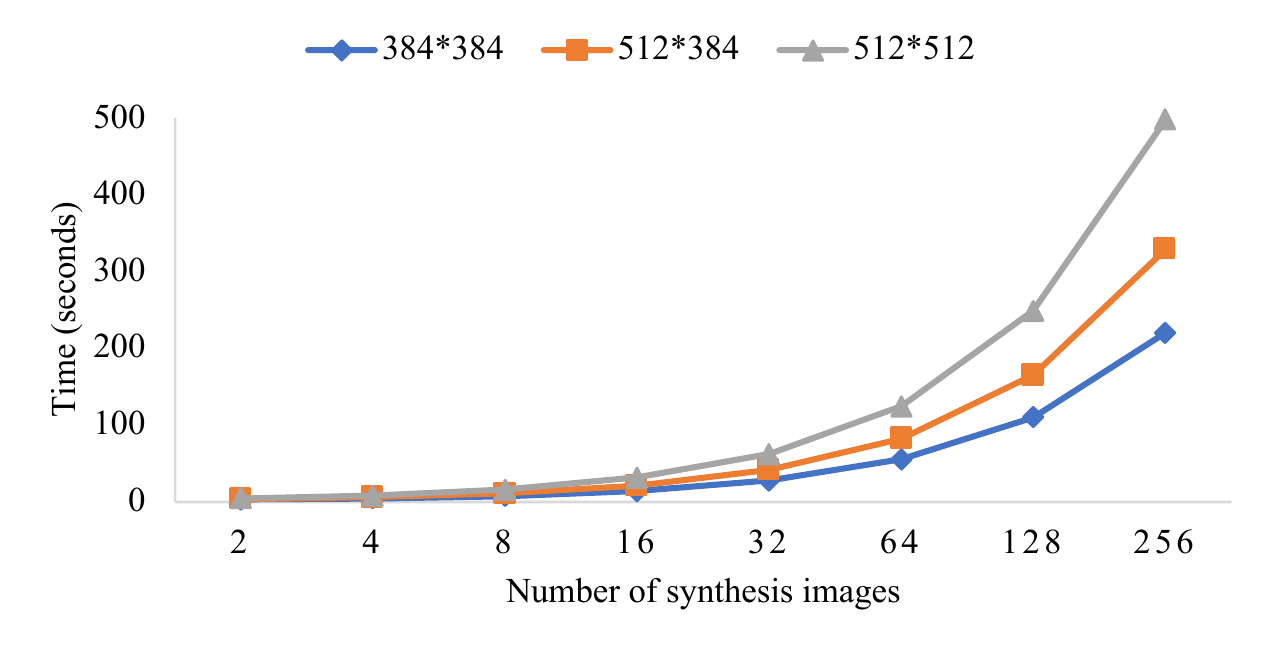}
  \caption{Stable Diffusion's execution time in different resolutions using a single Nvidia V100 GPU. The batch size is maximized for each respective resolution, and the hyperparameter of sampling steps is 20. The optimal numbers of synthesis images are from 8 to 16.}
  \label{fig:time_number_of_images_line_chart}
\end{figure}

\begin{figure}[t]
  \centering
  \includegraphics[width=0.85\linewidth]{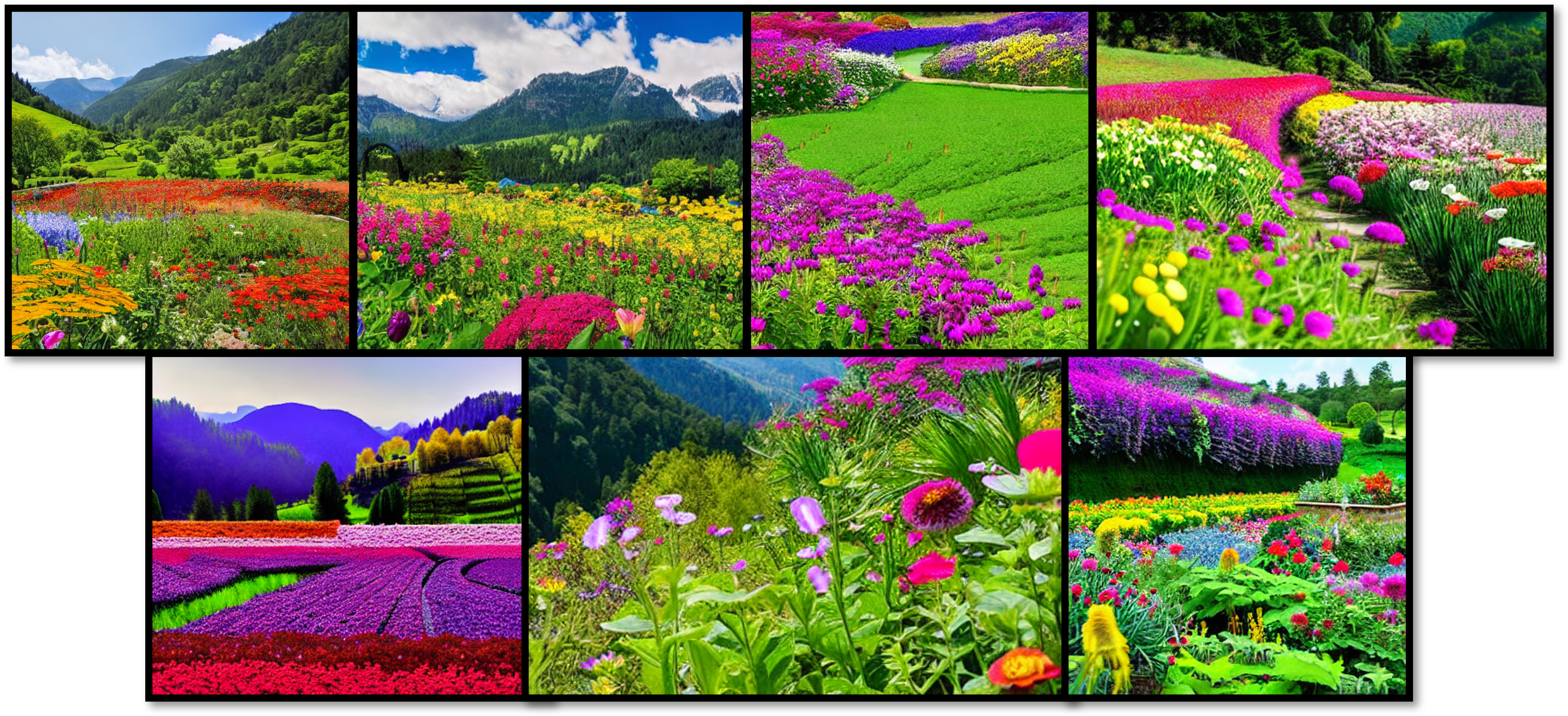}
  \caption{Images generated by Stable Diffusion using the same prompt "A beautiful flower garden \newcommentAdd{on} a sunny day with a valley background" in decreasing resolution order (from left to right, top to bottom).}
  \vspace{-2mm}
  \label{fig:grid}
\end{figure}

Image resolution is another element that significantly affects Stable Diffusion's running time. Following tips of Suraj~\etal~\cite{Suraj-huggingface2022}, we tried reducing the resolution and came up with seven different resolutions in decreasing execution time, namely $512\times 512$, $512\times 448$, $448\times 448$, $512\times 384$, $448\times 384$, $512\times 320$, and $384\times 384$. As illustrated in Fig. \ref{fig:grid}, the first two images on the top row have a valley background. As the resolution decreases, contextual information in the prompt will gradually become less constrained.

Fully utilizing GPU's capability is another technique to enhance the performance of the model. We tried setting the largest batch size on each respective resolution and recorded the run time accordingly. Figure~\ref{fig:time_number_of_images_line_chart} illustrates the benchmark result of the highest resolution, lowest resolution, and median one. The experimental result shows that the optimal number of synthesis images (\ie, K in Eq.~\ref{eq:similarity}) is either 8 or 16. The reason is twofold: Firstly, a small batch size can easily fit into a conventional GPU; Secondly, since the image captioning module's execution time scales linearly with the number of generated images, selecting a small batch size can thus improve both modules' performance.

\subsection{Image Captioning Visualization}

We employed GRIT \cite{Nguyen-arxiv2022} model that uses the pre-trained of object detector on four datasets: COCO \cite{Lin-ECCV2014}, Visual Genome, Open Images \cite{Kuznetsova-IJCV2020}, Object365 \cite{Shao-ICCV2019}\newcommentAdd{,} and applied Parallel Cross-Attention \cite{Nguyen-arxiv2022} for image captioning. We can achieve the per-batch inference time of about 0.75s when setting the batch size of 16 and 0.87s with the batch size of 8 on a single Nvidia V100 GPU.

\begin{figure}[t]
    \centering
    \includegraphics[width=1\linewidth]{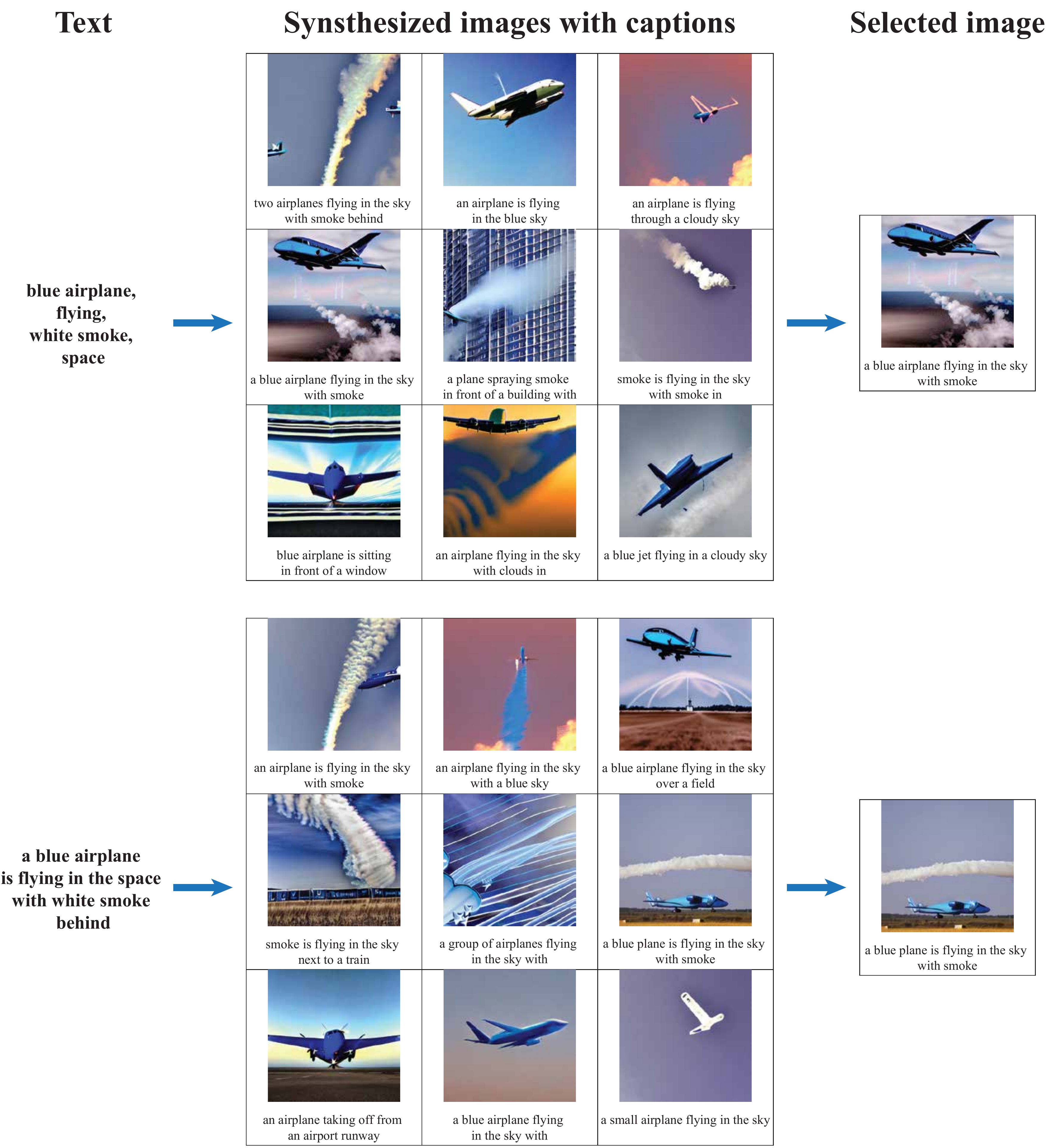}
    \caption{Example of text refinement on a complete sentence (below) and only on keywords (above) giving similar results.}
    \label{fig:example_image_captioning}
\end{figure}

Example results are visualized in Fig. \ref{fig:example_image_captioning}. With the developed text refinement mechanism, our MUGCAT system obviously generates an illustration and complete caption with high semantic similarity with the original sentence from the keywords, as shown in Fig.\ref{fig:example_image_captioning}. 

\section{Conclusion}
\label{sec:conclusion}

We have proposed a \textbf{MU}ltilin\textbf{G}ual \textbf{C}ommunic\textbf{AT}ion system (MUGCAT), which integrates sign language recognition and translation, text-to-image, and image captioning methods. The proposed system harmonizes three different methodologies to help overcome the difficulty of communicating with deaf individuals. Leveraging the latest development in text-to-image synthesis and image captioning to transform written text into visual images, we strive to lift the language barrier that has always existed in the sign language community. Experiments show the potential of our proposed system in practice. In the future, it would be interesting to modify our system's camera to first-person. We want to explore the possibility of sign language recognition and translation methods from a first-person perspective since this would overcome the problem of requiring standing in front of a fixed camera.


\textbf{Acknowledgment.} This research is funded by University of Science, VNU-HCM, under grant number CNTT 2022-15.

\balance

\tiny{
\bibliography{short_bibtex}
\bibliographystyle{IEEEtran}
}

\end{document}